\documentclass[10pt,twocolumn,letterpaper]{article}

\usepackage{cvpr}
\usepackage{times}
\usepackage{epsfig}
\usepackage{graphicx}
\usepackage{amsmath}
\usepackage{amssymb}

\usepackage[breaklinks=true,bookmarks=false,colorlinks]{hyperref}

\cvprfinalcopy 

\def\cvprPaperID{1128} 

\usepackage{booktabs}
\usepackage{multirow}
\usepackage{graphicx}
\usepackage{subcaption}
\usepackage{mmstyle}

\ifcvprfinal\fi
\begin{document}
\title{Libra R-CNN: Towards Balanced Learning for Object Detection}
\author{
Jiangmiao Pang$^{\dag}$~~~~~
Kai Chen$^{\S}$~~~~~
Jianping Shi$^{\ddagger}$~~~~~
Huajun Feng$^{\dag}$~~~~~
Wanli Ouyang$^{\flat}$~~~~~
Dahua Lin$^{\S}$
\\
\and
$^{\dag}$Zhejiang University~~~~~~$^{\S}$The Chinese University of Hong Kong\\
$^{\ddagger}$SenseTime Research~~~~~~$^{\flat}$The University of Sydney\\
\and
{\tt\small pjm@zju.edu.cn~~~ck015@ie.cuhk.edu.hk~~~shijianping@sensetime.com}
\and
{\tt\small fenghj@zju.edu.cn~~~wanli.ouyang@sydney.edu.au~~~dhlin@ie.cuhk.edu.hk}
}
\maketitle
\thispagestyle{empty}


\begin{abstract}
Compared with model architectures, the training process, which is also
crucial to the success of detectors,
has received relatively less attention in object detection.
In this work, we carefully revisit the standard training practice of detectors,
and find that the detection performance is often limited by the imbalance
during the training process, which generally consists in three levels --
sample level, feature level, and objective level.
To mitigate the adverse effects caused thereby, we propose
Libra R-CNN, a simple but effective framework towards balanced learning for object detection.
It integrates three novel components:
IoU-balanced sampling, balanced feature pyramid, and balanced L1 loss,
respectively for reducing the imbalance at sample, feature, and objective level.
Benefitted from the overall balanced design, Libra R-CNN significantly improves the detection performance.
Without bells and whistles, it achieves 2.5 points and 2.0 points higher Average Precision (AP) than FPN
Faster R-CNN and RetinaNet respectively on MSCOCO.
\footnote{Code is available at \url{https://github.com/OceanPang/Libra_R-CNN}.}
\end{abstract}


\section{Introduction}

Along with the advances in deep convolutional networks, recent years have
seen remarkable progress in object detection.
A number of detection frameworks such as
Faster R-CNN~\cite{frcnn},
RetinaNet~\cite{focalloss},
and Cascaded R-CNN~\cite{cascadercnn} have been developed,
which have substantially pushed forward the state of the art.
Despite the apparent differences in the pipeline architectures,
\eg~single-stage \vs two-stage, modern detection frameworks mostly follow
a common training paradigm, namely, sampling regions, extracting features therefrom,
and then jointly recognizing the categories and refining the locations
under the guidance of a standard multi-task objective function.

\begin{figure}
	\setlength{\belowcaptionskip}{-15pt}
	\centering
	\includegraphics[width=\linewidth]{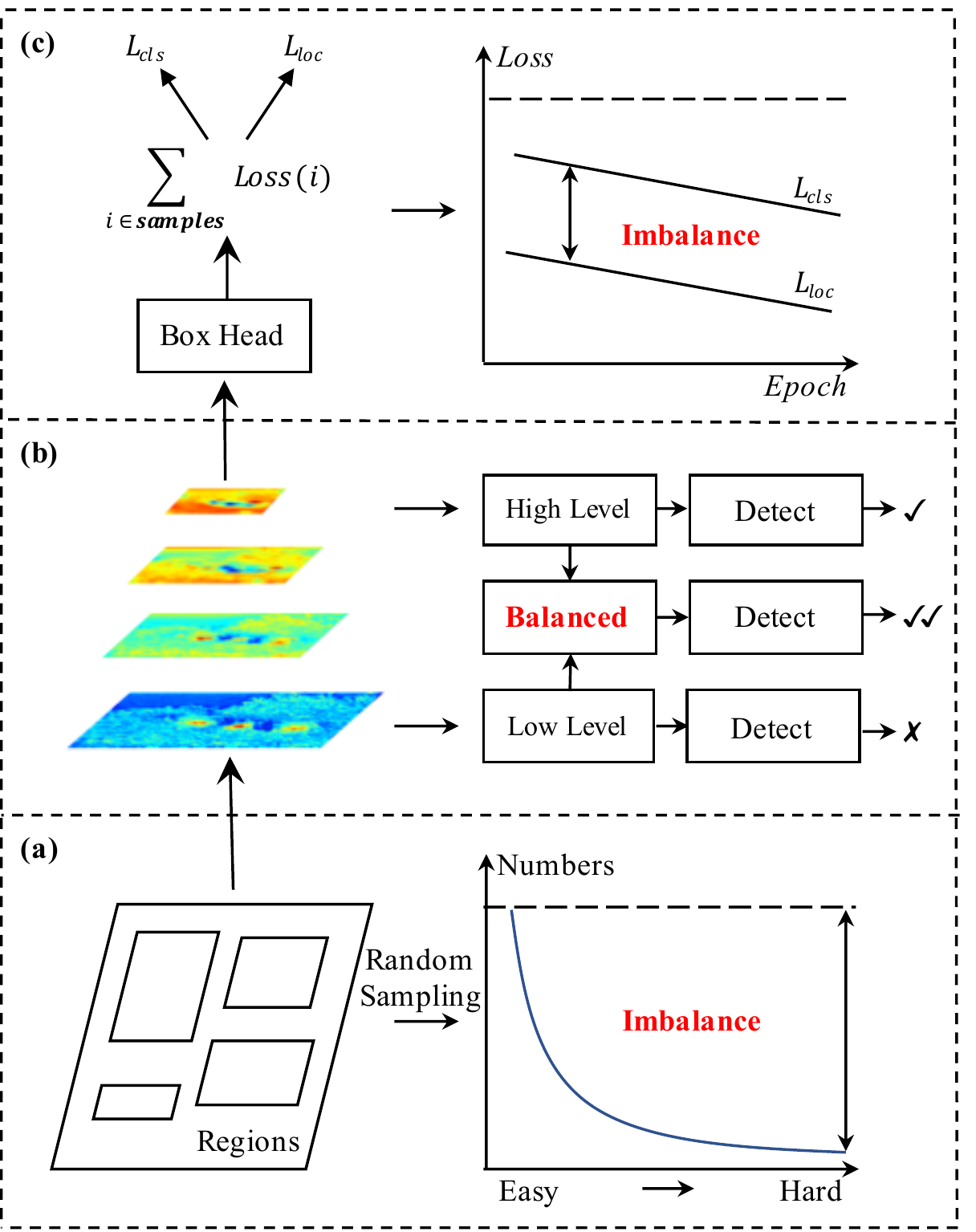}
	\caption{Imbalance consists in (a) sample level (b) feature level and (c) objective level,
	which prevents the well-designed model architectures from being fully exploited.}
	\label{fig:imbalance}
\end{figure}

Based on this paradigm, the success of the object detector training depends on
three key aspects:
(1) whether the selected region samples are representative,
(2) whether the extracted visual features are fully utilized, and
(3) whether the designed objective function is optimal.
However, our study reveals that the typical training process is significantly imbalanced in all these aspects.
This imbalance issue prevents the power of well-designed model architectures from being fully exploited,
thus limiting the overall performance, which is shown in Figure~\ref{fig:imbalance}.
Below, we describe these issues in turn:

\vspace{-15pt}
\paragraph{\emph{1) Sample level imbalance:}}
When training an object detector, hard samples are particularly valuable
as they are more effective to improve the detection performance.
However, the random sampling scheme usually results in
the selected samples dominated by easy ones.
The popularized hard mining methods, \eg~OHEM~\cite{ohem}, can help driving
the focus towards hard samples.
However, they are often sensitive to noise labels and incurring considerable memory and computing costs.
Focal loss~\cite{focalloss} also alleviates this problem in single-stage detectors,
but is found little improvement when extended to R-CNN as the majority easy negatives are filtered by the two-stage procedure.
Hence, this issue needs to be solved more elegantly.

\vspace{-15pt}
\paragraph{\emph{2) Feature level imbalance:}}
Deep high-level features in backbones are with more semantic meanings while the shallow low-level features are more content descriptive~\cite{zeiler2014visualizing}.
Recently, feature integration via lateral connections in FPN~\cite{fpn} and PANet~\cite{panet} have advanced the development of object detection.
These methods inspire us that the low-level and high-level information are complementary for object detection.
The approach that how them are utilized to integrate the pyramidal representations determines the detection performance.
However, what is the best approach to integrate them together?
Our study reveals that the integrated features should possess balanced information from each resolution.
But the sequential manner in aforementioned methods will make
integrated features focus more on adjacent resolution but less on others.
The semantic information contained in non-adjacent levels would be diluted once per fusion during the information flow.

\vspace{-15pt}
\paragraph{\emph{3) Objective level imbalance:}}
A detector needs to carry out two tasks, \ie~classification and localization.
Thus two different goals are incorporated in the training objective.
If they are not properly balanced, one goal may be compromised, leading to suboptimal performance overall~\cite{kendall2017multi}.
The case is the same for the involved samples during the training process.
If they are not properly balanced, the small gradients produced by the easy samples may be drowned into the large gradients produced by the hard ones, thus limiting further refinement.
Hence, we need to rebalance the involved tasks and samples towards the optimal convergence.


To mitigate the adverse effects caused by these issues,
we propose Libra R-CNN, a simple but effective framework for object detection that
explicitly enforces the balance at all three levels discussed above.
This framework integrates three novel components:
(1) \emph{IoU-balanced sampling}, which mines hard samples according to their IoU with assigned ground-truth.
(2) \emph{balanced feature pyramid},
which strengthens the multi-level features using the same deeply integrated balanced semantic features.
(3) \emph{balanced L1 loss}, which promotes crucial gradients, to rebalance the involved classification, overall localization and accurate localization.

Without bells and whistles, Libra R-CNN
achieves 2.5 points and 2.0 points higher Average Precision (AP) than FPN Faster R-CNN and
RetinaNet respectively on MS COCO~\cite{coco}.
With the 1$\times$ schedule in~\cite{Detectron2018},
Libra R-CNN can obtain 38.7 and 43.0 AP with FPN Faster R-CNN based on ResNet-50
and ResNeXt-101-64x4d respectively.

Here, we summarize our main contributions:
(1) We systematically revisit the training process of detectors.
Our study reveals the imbalance problems at three levels that limit the
detection performance.
(2) We propose Libra R-CNN, a framework that rebalances the training
process by combining three new components: IoU-balanced sampling,
balanced feature pyramid, and balanced L1 loss.
(3) We test the proposed framework on MS COCO, consistently obtaining
significant improvements over state-of-the-art detectors, including
both single-stage and two-stage ones.


\section{Related Work}

\begin{figure*}
	\centering
	\includegraphics[width=\linewidth]{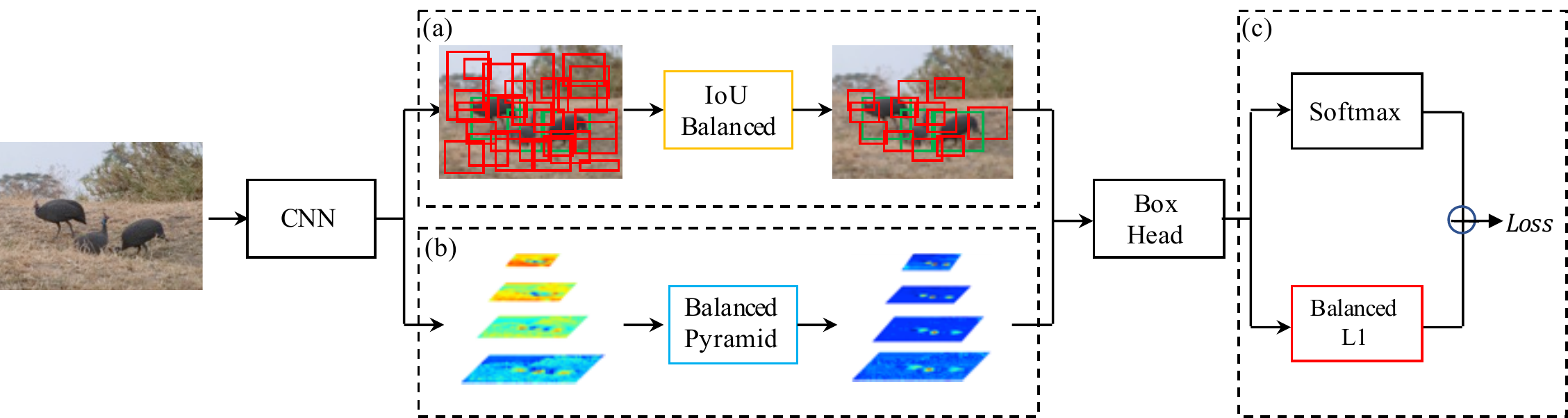}
	\caption{Overview of the proposed Libra R-CNN:
		an overall balanced design for object detection which integrated three novel components (a) IoU-balanced sampling (b) balanced feature pyramid and (c) balanced L1 loss, respectively for reducing the imbalance at sample, feature, and objective level.}
	\label{fig:overall}
\end{figure*}

\vspace{0pt}
\paragraph{Model architectures for object detection.}
Recently, object detection are popularized by both two-stage and single-stage detectors.
Two-stage detectors were first introduced by R-CNN~\cite{rcnn}.
Gradually derived SPP~\cite{spp}, Fast R-CNN~\cite{fastrcnn} and Faster R-CNN~\cite{frcnn} promoted the developments furthermore.
Faster R-CNN proposed region proposal network to improve the efficiency of detectors and allow the detectors to be trained end-to-end.
After this meaningful milestone, lots of methods were introduced to enhance Faster R-CNN from different points.
For example,
FPN~\cite{fpn} tackled the scale variance via pyramidal predictions.
Cascade R-CNN~\cite{cascadercnn} extended Faster R-CNN to a multi-stage detector through the classic yet powerful cascade architecture.
Mask R-CNN~\cite{maskrcnn} extended Faster R-CNN by adding a mask branch that refines the detection results under the help of multi-task learning.
HTC~\cite{htc} further improved the mask information flow in Mask R-CNN through a new cascade architecture.
On the other hand, single-stage detectors are popularized by YOLO~\cite{yolo, yolo9000} and SSD~\cite{ssd}.
They are simpler and faster than two-stage detectors but have trailed the accuracy until the introduction of RetinaNet~\cite{focalloss}.
CornetNet~\cite{cornernet} introduced an insight that the bounding boxes can be predicted as a pair of key points.
Other methods focus on
	cascade procedures~\cite{ouyang2017chained},
	duplicate removal~\cite{relationnetwork, learningnms},
	multi-scales~\cite{cai2016unified, bell2016inside, sniper2018, snip},
	adversarial learning~\cite{adapt}
and more contextual~\cite{zeng2018crafting}.
All of them made significant progress from different concerns.

\vspace{-12pt}
\paragraph{Balanced learning for object detection.}
Alleviating imbalance in the training process of object detection is crucial to achieve an optimal training and fully exploit the potential of model architectures.

\vspace{-14pt}
\paragraph{\emph{{Sample level imbalance.}}}
OHEM~\cite{ohem} and focal loss~\cite{focalloss} are primary existing solutions for sample level imbalance in object detection.
The commonly used OHEM automatically selects hard samples according to their confidences.
However, this procedure causes extra memory and speed costs, making the training process bloated.
Moreover, the OHEM also suffers from noise labels so that it cannot work well in all cases.
Focal loss solved the extra foreground-background class imbalance in single-stage detectors with an elegant loss formulation,
but it generally brings little or no gain to two-stage detectors because of the different imbalanced situation.
Compared with these methods, our method is substantially lower cost, and tackles the problem elegantly.

\vspace{-14pt}
\paragraph{\emph{{Feature level imbalance.}}}
Utilizing multi-level features to generate discriminative pyramidal representations is crucial to detection performance.
FPN~\cite{fpn} proposed lateral connections to enrich the semantic information of shallow layers through a top-down pathway.
After that, PANet~\cite{panet} brought in a bottom-up pathway to further increase the low-level information in deep layers.
Kong \etal~\cite{kong2018deep} proposed a novel efficient pyramid based on SSD that integrates the features in a highly-nonlinear yet efficient way.
Different from these methods, our approach relies on integrated balanced semantic features to strengthen original features.
In this manner, each resolution in the pyramid obtains equal information from others, thus balancing the information flow and leading the features more discriminative.

\vspace{-14pt}
\paragraph{\emph{{Objective level imbalance.}}}
Kendall \etal ~\cite{kendall2017multi} had proved that the performance of models based on multi-task learning is strongly dependent on the relative weight between the loss of each task.
But previous approaches~\cite{frcnn, fpn, focalloss} mainly focus on how to enhance the recognition ability of model architectures.
Recently, UnitBox~\cite{unitbox} and IoU-Net~\cite{iounet} introduced some new objective functions related to IoU, to promote the localization accuracy.
Different to them, our method rebalances the involved tasks and samples to achieve a better convergence.

\begin{figure}
	\centering
	\includegraphics[width=\linewidth]{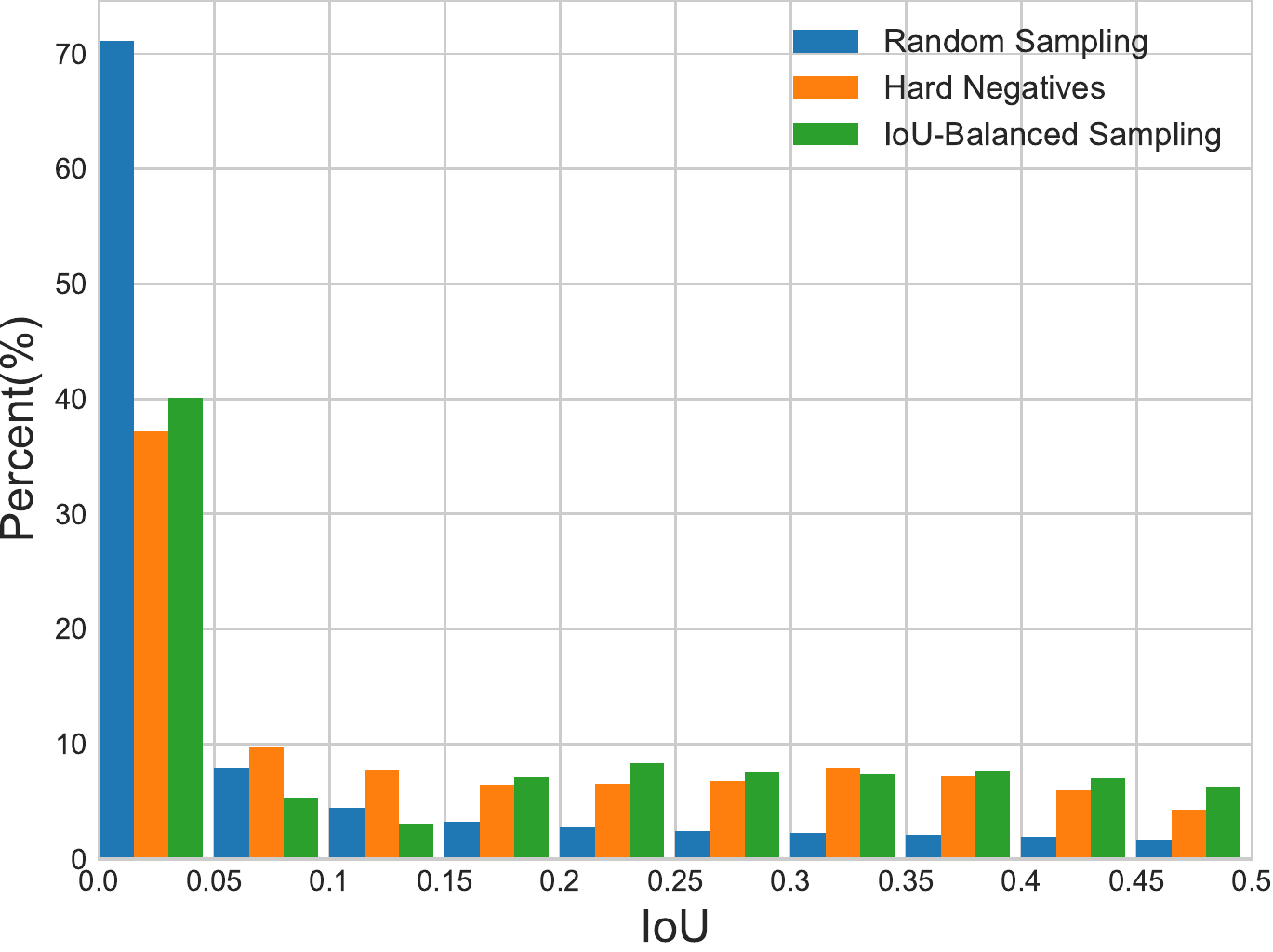}
	\caption{IoU distribution of random selected samples, IoU-balanced selected samples, and hard negatives.}
	\label{fig:sample_dist}
\end{figure}

\section{Methodology}

The overall pipeline of Libra R-CNN is shown in Figure~\ref{fig:overall}.
Our goal is to alleviate the imbalance exists in the training process of detectors using an overall balanced design,
thus exploiting the potential of model architectures as much as possible.
All components will be detailed in the following sections.

\begin{figure*}
	\centering
	\includegraphics[width=\linewidth]{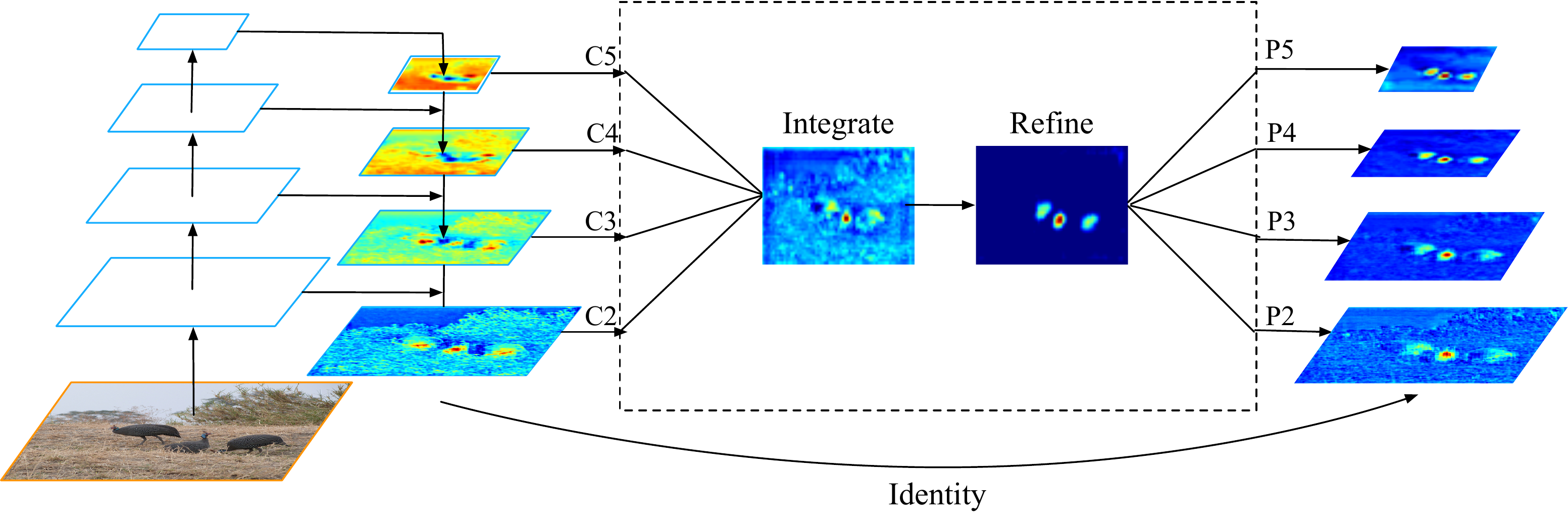}
	\caption{Pipeline and heatmap visualization of balanced feature pyramid.}
	\label{fig:fpn}
\end{figure*}

\subsection{IoU-balanced Sampling}
\label{subsec:sample}
Let us start with the basic question:
is the overlap between a training sample and its corresponding ground truth associated with its difficulty?
To answer this question, we conduct experiments to find the truth behind.
Results are shown in Figure~\ref{fig:sample_dist}.
We mainly consider hard negative samples, which are known to be the main problem.
We find that more than $60\%$ hard negatives have an overlap greater than $0.05$,
but random sampling only provides us $30\%$ training samples that are greater than the same threshold.
This extreme sample imbalance buries many hard samples into thousands of easy samples.

Motivated by this observation, we propose IoU-balanced sampling: a simple but effective hard mining method without extra cost.
Suppose we need to sample $N$ negative samples from $M$ corresponding candidates.
The selected probability for each sample under random sampling is
\begin{equation}
	\label{equ:random}
	p = \frac{N}{M}.
\end{equation}

To raise the selected probability of hard negatives,
we evenly split the sampling interval into $K$ bins according to IoU.
$N$ demanded negative samples are equally distributed to each bin.
Then we select samples from them uniformly.
Therefore, we get the selected probability under IoU-balanced sampling
\begin{equation}
	\label{equ:sample}
	p_{k} = \frac{N}{K} *\frac{1}{M_{k}}, ~~ k \in [0, K),
\end{equation}
where $M_{k}$ is the number of sampling candidates in the corresponding interval denoted by k.
K is set to 3 by default in our experiments.

The sampled histogram with IoU-balanced sampling is shown by green color in Figure~\ref{fig:sample_dist}.
It can be seen that our IoU-balanced sampling can guide the distribution of training samples close to the one of hard negatives.
Experiments also show that the performance is not sensitive to K,
as long as the samples with higher IoU are more likely selected.

Besides, it is also worth noting that the method is also suitable for hard positive samples.
However, in most cases, there are not enough sampling candidates to extend this procedure into positive samples.
To make the balanced sampling procedure more comprehensive,
we sample equal positive samples for each ground truth as an alternative method.

\begin{figure*}
	\centering
	\includegraphics[width=\linewidth]{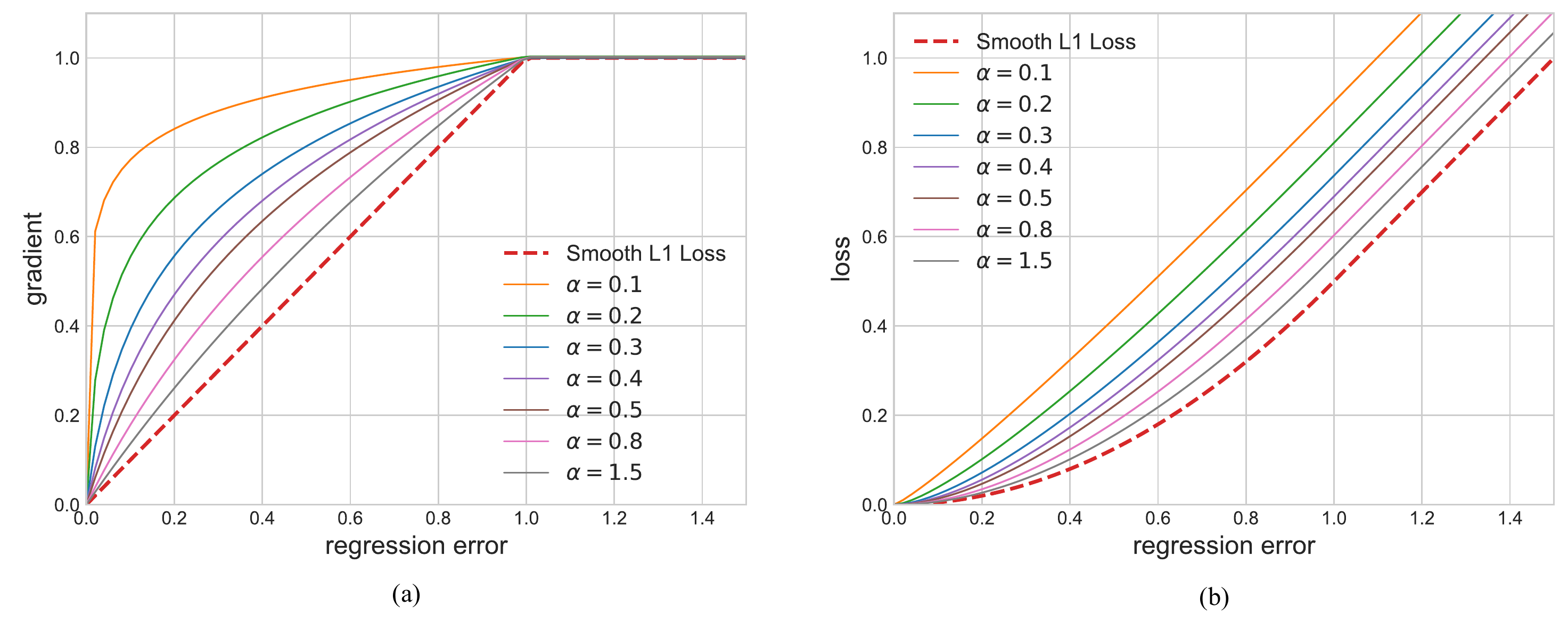}
	\caption{We show curves for (a) gradient and (b) loss of our balanced L1 loss here. Smooth L1 loss is also shown in dashed lines. $\gamma$ is set default as 1.0.}
	\label{fig:loss}
	\vspace{-10pt}
\end{figure*}

\subsection{Balanced Feature Pyramid}
Different from former approaches\cite{fpn,panet} that integrate multi-level features using lateral connections,
our key idea is to \emph{strengthen} the multi-level features using the \emph{same} deeply integrated balanced semantic features.
The pipeline is shown in Figure~\ref{fig:fpn}.
It consists of four steps, rescaling, integrating, refining and strengthening.

\vspace{-12pt}
\paragraph{Obtaining balanced semantic features.}
Features at resolution level $l$ are denoted as $C_l$.
The number of multi-level features is denoted as L.
The indexes of involved lowest and highest levels are denoted as $l_{min}$ and $l_{max}$.
In Figure~\ref{fig:fpn}, $C_2$ has the highest resolution.
To integrate multi-level features and preserve their semantic hierarchy at the same time, we first resize the multi-level features $\{C_{2}, C_{3}, C_{4}, C_{5}\}$ to an intermediate size, \ie, the same size as $C_{4}$, with interpolation and max-pooling respectively.
Once the features are rescaled, the \emph{balanced semantic features} are obtained by simple averaging as
\begin{equation}
	\label{equ:IDFeat}
	C = \frac{1}{L}\sum_{l=l_{min}}^{l_{max}}{C_{l}}.
\end{equation}
The obtained features are then rescaled using the same but reverse procedure to strengthen the original features.
Each resolution obtains equal information from others in this procedure.
Note that this procedure does not contain any parameter.
We observe improvement with this nonparametric method, proving the effectiveness of the information flow.

\vspace{-12pt}
\paragraph{Refining balanced semantic features.}
The balanced semantic features can be further refined to be more discriminative.
We found both the refinements with convolutions directly and the non-local module~\cite{nonlocal} work well.
But the non-local module works more stable.
Therefore, we use the embedded Gaussian non-local attention as default in this paper.
The refining step helps us enhance the integrated features and further improve the results.

With this method, features from low-level to high-level are aggregated at the same time.
The outputs $\{P_{2}, P_{3}, P_{4}, P_{5}\}$ are used for object detection following the same pipeline in FPN.
It is also worth mentioning that our balanced feature pyramid can work as complementary with recent solutions such as FPN and PAFPN without any conflict.

\begin{table*}[htb]
	\centering
	\caption{Comparisons with state-of-the-art methods on COCO \emph{test-dev}.
	  The symbol ``*'' means our re-implemented results.
	  The ``$1\times$'', ``$2\times$'' training schedules follow the settings explained in Detectron~\cite{Detectron2018}.}
	\vspace{-5pt}
	\addtolength{\tabcolsep}{1pt}
	\begin{tabular}{*{12}{c}}
		\toprule
		Method                       & Backbone  & Schedule   &  AP & $\text{AP}_{50}$ & $\text{AP}_{75}$ & $\text{AP}_{S}$ & $\text{AP}_{M}$ & $\text{AP}_{L}$  \\
		\midrule
		YOLOv2~\cite{yolo9000} & DarkNet-19 & - & 21.6 & 44.0 & 19.2 & 5.0 & 22.4 & 35.5\\
		SSD512~\cite{ssd} & ResNet-101 & - & 31.2 & 50.4 & 33.3 & 10.2 & 34.5 & 49.8 \\
		RetinaNet~\cite{focalloss} & ResNet-101-FPN & -  & 39.1 & 59.1 & 42.3 & 21.8 & 42.7 & 50.2 \\
		Faster R-CNN~\cite{fpn} & ResNet-101-FPN & - & 36.2 & 59.1 & 39.0 & 18.2 & 39.0 & 48.2 \\
		Deformable R-FCN~\cite{rfcn} & Inception-ResNet-v2 & -  & 37.5 & 58.0 & 40.8 & 19.4 & 40.1 & 52.5 \\
		Mask R-CNN~\cite{maskrcnn} & ResNet-101-FPN & - & 38.2 & 60.3 & 41.7 & 20.1 & 41.1 & 50.2 \\
		\midrule
		Faster R-CNN$^*$ & ResNet-50-FPN & $1\times$ & 36.2 & 58.5 & 38.9 & 21.0 & 38.9 & 45.3            \\
		Faster R-CNN$^*$ & ResNet-101-FPN & $1\times$ & 38.8 & 60.9 & 42.1 & 22.6 & 42.4 & 48.5         \\
		Faster R-CNN$^*$ & ResNet-101-FPN & $2\times$ & 39.7 & 61.3 & 43.4 & 22.1 & 43.1 & 50.3           \\
		Faster R-CNN$^*$ & ResNeXt-101-FPN & $1\times$ & 41.9 & 63.9 & 45.9 & 25.0 & 45.3 & 52.3 \\
		RetinaNet$^*$ & ResNet-50-FPN & $1\times$ & 35.8 & 55.3 & 38.6 & 20.0 & 39.0 & 45.1  \\
		\midrule
		Libra R-CNN (ours) & ResNet-50-FPN & $1\times$ & 38.7 & 59.9 & 42.0 & 22.5 & 41.1 & 48.7 \\
		Libra R-CNN (ours) & ResNet-101-FPN & $1\times$ & 40.3 & 61.3 & 43.9 & 22.9 & 43.1 & 51.0 \\
		Libra R-CNN (ours) & ResNet-101-FPN & $2\times$ & 41.1 & 62.1 & 44.7 & 23.4 & 43.7 & 52.5 \\
		Libra R-CNN (ours) & ResNeXt-101-FPN  & $1\times$ & 43.0 & 64.0 & 47.0 & 25.3 & 45.6 & 54.6 \\
		Libra RetinaNet (ours) & ResNet-50-FPN & $1\times$ & 37.8 & 56.9 & 40.5 & 21.2 & 40.9 & 47.7  \\
		\bottomrule
	\end{tabular}
	\vspace{-5pt}
	\label{tab:overall-results}
\end{table*}

\subsection{Balanced L1 Loss}
Classification and localization problems are solved simultaneously under the guidance of a multi-task loss since Fast R-CNN~\cite{fastrcnn}, which is defined as
\begin{equation}
	\label{equ:multi-task}
	L_{p, u, t^u, v} = L_{cls}(p, u) + \lambda[u \ge 1]L_{loc}(t^u, v).
\end{equation}
$L_{cls}$ and $L_{loc}$ are objective functions corresponding to recognition and localization respectively.
Predictions and targets in $L_{cls}$ are denoted as $p$ and $u$.
$t^u$ is the corresponding regression results with class $u$.
$v$ is the regression target.
$\lambda$ is used for tuning the loss weight under multi-task learning.
We call samples with a loss greater than or equal to 1.0 outliers.
The other samples are called inliers.

A natural solution for balancing the involved tasks is to tune the loss weights of them.
However, owing to the unbounded regression targets,
directly raising the weight of localization loss will make the model more sensitive to outliers.
These outliers, which can be regarded as hard samples, will produce excessively large gradients that are harmful to the training process.
The inliers, which can be regarded as the easy samples, contribute little gradient to the overall gradients compared with the outliers.
To be more specific, inliers only contribute 30\% gradients average per sample compared with outliers.
Considering these issues, we propose balanced L1 loss, which is denoted as $L_{b}$.

Balanced L1 loss is derived from the conventional smooth L1 loss,
in which an inflection point is set to separate inliers from outliners, and clip the large gradients produced by outliers with a maximum value of 1.0, as shown by the dashed lines in Figure~\ref{fig:loss}-(a).
The key idea of balanced L1 loss is promoting the crucial regression gradients,
\ie gradients from inliers (accurate samples), to rebalance the involved samples and tasks, thus achieving a more balanced training within classification, overall localization and accurate localization.
Localization loss $L_{loc}$ uses balanced L1 loss is defined as
\begin{equation}
	\label{equ:loc}
	L_{loc} = \sum_{i\in\{x,y,w,h\}}{L_{b}(t_{i}^{u} - v_i)},
\end{equation}
and its corresponding formulation of gradients follows
\begin{equation}
	\label{equ:partial}
	\frac{\partial{L_{loc}}}{\partial{w}} \varpropto
	\frac{\partial{L_{b}}}{\partial{t_i^u}} \varpropto
	\frac{\partial{L_{b}}}{\partial{x}},
\end{equation}
Based on the formulation above, we design a promoted gradient formulation as
\begin{equation}
	\label{equ:gradient}
	\frac{\partial{L_{b}}}{\partial{x}} =
	\begin{cases}
		\alpha ln(b|x| + 1) & \text{if $|x| < 1$} \\
		\gamma  & \text{otherwise},
	\end{cases}
\end{equation}

Figure~\ref{fig:loss}-(a) shows that our balanced L1 loss increases the gradients of inliers under the control of a factor denoted as $\alpha$.
A small $\alpha$ increases more gradient for inliers, but the gradients of outliers are not influenced.
Besides, an overall promotion magnification controlled by $\gamma$ is also brought in for tuning the upper bound of regression errors, which can help the objective function better balancing involved tasks.
The two factors that control different aspects are mutually enhanced to reach a more balanced training.
$b$ is used to ensure $L_{b}(x=1)$ has the same value for both formulations in Eq. (\ref{equ:loss}).

By integrating the gradient formulation above, we can get the balanced L1 loss
\begin{equation}
	\label{equ:loss}
	L_{b}(x) =
	\begin{cases}
		\frac{\alpha}{b}(b|x| + 1)ln(b|x| + 1) - \alpha |x|  & \text{if $|x| < 1$}\\
		\gamma|x| + C  & \text{otherwise},
	\end{cases}
\end{equation}
in which the parameters $\gamma$, $\alpha$, and $b$ are constrained by
\begin{equation}
	\alpha ln(b+1) = \gamma.
\end{equation}
The default parameters are set as $\alpha = 0.5$ and $\gamma = 1.5$ in our experiments.


\section{Experiments}
\subsection{Dataset and Evaluation Metrics}
All experiments are implemented on the challenging MS COCO~\cite{coco} dataset.
It consists of 115k images for training (\emph{train-2017}) and 5k images for validation (\emph{val-2017}).
There are also 20k images in \emph{test-dev} that have no disclosed labels.
We train models on \emph{train-2017}, and report ablation studies and final results on \emph{val-2017} and \emph{test-dev} respectively.
All reported results follow standard COCO-style Average Precision (AP) metrics that include
AP (averaged over IoU thresholds), AP$_{50}$ (AP for IoU threshold 50\%), AP$_{75}$ (AP for IoU threshold 75\%).
We also include AP$_{S}$, AP$_M$, AP$_L$, which correspond to the results on small, medium and large scales respectively.
The COCO-style Average Recall (AR) with AR$^{100}$, AR$^{300}$, AR$^{1000}$ correspond to the average recall when there are 100, 300 and 1000 proposals per image respectively.

\subsection{Implementation Details}
For fair comparisons, all experiments are implemented on PyTorch~\cite{pytorch} and mmdetection~\cite{mmdetection2018}.
The backbones used in our experiments are publicly available.
We train detectors with 8 GPUs (2 images per GPU) for 12 epochs with an initial learning rate of 0.02, and decrease it by 0.1 after 8 and 11 epochs respectively if not specifically noted.
All other hyper-parameters follow the settings in mmdetection~\cite{mmdetection2018} if not specifically noted.

\begin{table*}[htb]
	\centering
	\caption{Effects of each component in our Libra R-CNN. Results are reported on COCO \emph{val-2017}.}
	\vspace{0.1cm}
	\addtolength{\tabcolsep}{0pt}
	\begin{tabular}{*{12}{c}}
		\toprule
		IoU-balanced Sampling & Balanced Feature Pyramid & Balanced L1 Loss & AP   & $\text{AP}_{50}$ & $\text{AP}_{75}$ & $\text{AP}_{S}$ & $\text{AP}_{M}$ & $\text{AP}_{L}$ \\
		\midrule
		                      &                          &                  & 35.9 & 58.0             & 38.4             & 21.2            & 39.5            & 46.4            \\
		\checkmark            &                          &                  & 36.8 & 58.0             & 40.0             & 21.1            & 40.3            & 48.2            \\
		\checkmark            & \checkmark               &                  & 37.7 & 59.4             & 40.9             & 22.4            & 41.3            & 49.3            \\
		\checkmark            & \checkmark               & \checkmark       & 38.5 & 59.3             & 42.0             & 22.9            & 42.1            & 50.5            \\
		\bottomrule
	\end{tabular}
	\label{tab:overall-ablation}
\end{table*}

\begin{table}[htb]
	\centering
	\caption{Comparisons between Libra RPN and RPN. The symbol ``*'' means our re-implements.}
	\vspace{-2pt}
	\addtolength{\tabcolsep}{-5pt}
	\begin{tabular}{*{12}{c}}
		\toprule
		Method           & Backbone        & $\text{AR}^{100}$ & $\text{AR}^{300}$ & $\text{AR}^{1000}$ \\
		\midrule
		RPN$^*$          & ResNet-50-FPN   & 42.5              & 51.2              & 57.1               \\
		RPN$^*$          & ResNet-101-FPN  & 45.4              & 53.2              & 58.7               \\
		RPN$^*$          & ResNeXt-101-FPN & 47.8              & 55.0              & 59.8               \\
		\midrule
		Libra RPN (ours) & ResNet-50-FPN   & 52.1              & 58.3              & 62.5               \\
		\bottomrule
	\end{tabular}
	\label{tab:rpn}
\end{table}

\begin{table}[htb]
	\centering
	\caption{Ablation studies of IoU-balanced sampling on COCO \emph{val-2017}.}
	\vspace{-5pt}
	\addtolength{\tabcolsep}{-2pt}
	\begin{tabular}{*{12}{c}}
		\toprule
		Settings    & AP   & $\text{AP}_{50}$ & $\text{AP}_{75}$ & $\text{AP}_{S}$ & $\text{AP}_{M}$ & $\text{AP}_{L}$ \\
		\midrule
		Baseline    & 35.9 & 58.0             & 38.4             & 21.2            & 39.5            & 46.4            \\
		\midrule
		Pos Balance & 36.1 & 58.2             & 38.2             & 21.3            & 40.2            & 47.3            \\
		$K = 2$     & 36.7 & 57.8             & 39.9             & 20.5            & 39.9            & 48.9            \\
		$K = 3$     & 36.8 & 57.9             & 39.8             & 21.4            & 39.9            & 48.7            \\
		$K = 5$     & 36.7 & 57.7             & 39.9             & 19.9            & 40.1            & 48.7            \\
		\bottomrule
	\end{tabular}
	\vspace{-10pt}
	\label{tab:iou}
\end{table}

\subsection{Main Results}
We compare Libra R-CNN with the state-of-the-art object detection approaches on the COCO \emph{test-dev} in Tabel~\ref{tab:overall-results}.
For fair comparisons with corresponding baselines, we report our re-implemented results of them, which are generally higher than that were reported in papers.
Through the overall balanced design, Libra R-CNN achieves 38.7 AP with ResNet-50~\cite{resnet}, which is $2.5$ points higher AP than FPN Faster R-CNN.
With ResNeXt-101-64x4d~\cite{resnext}, a much more powerful feature extractor, Libra R-CNN achieves 43.0 AP.

Apart from the two-stage frameworks, we further extend our Libra R-CNN to single stage detectors and report the results of Libra RetinaNet.
Considering that there is no sampling procedure in RetinaNet~\cite{focalloss}, Libra RetinaNet only integrates balanced feature pyramid and balanced L1 loss.
Without bells and whistles, Libra RetinaNet brings 2.0 points higher AP with ResNet-50 and achieves 37.8 AP.

Our method can also enhance the average recall of proposal generation.
As shown in Table~\ref{tab:rpn}, Libra RPN brings $9.2$ points higher AR$^{100}$, $6.9$ points higher AR$^{300}$ and $5.4$ points higher AR$^{1000}$ compared with RPN with ResNet-50 respectively.
Note that larger backbones only bring little gain to RPN.
Libra RPN can achieve 4.3 points higher AR$^{100}$ than ResNeXt-101-64x4d only with a ResNet-50 backbone.
The significant improvements from Libra RPN validate that the potential of RPN is much more exploited with the effective balanced training.

\subsection{Ablation Experiments}
\paragraph{Overall Ablation Studies.}
To analyze the importance of each proposed component, we report the overall ablation studies in Table~\ref{tab:overall-ablation}.
We gradually add IoU-balanced sampling, balanced feature pyramid and balanced L1 loss on ResNet-50 FPN Faster R-CNN baseline.
Experiments for ablation studies are implemented with the same pre-computed proposals for fair comparisons.

\begin{figure}
	\centering
	\includegraphics[width=\linewidth]{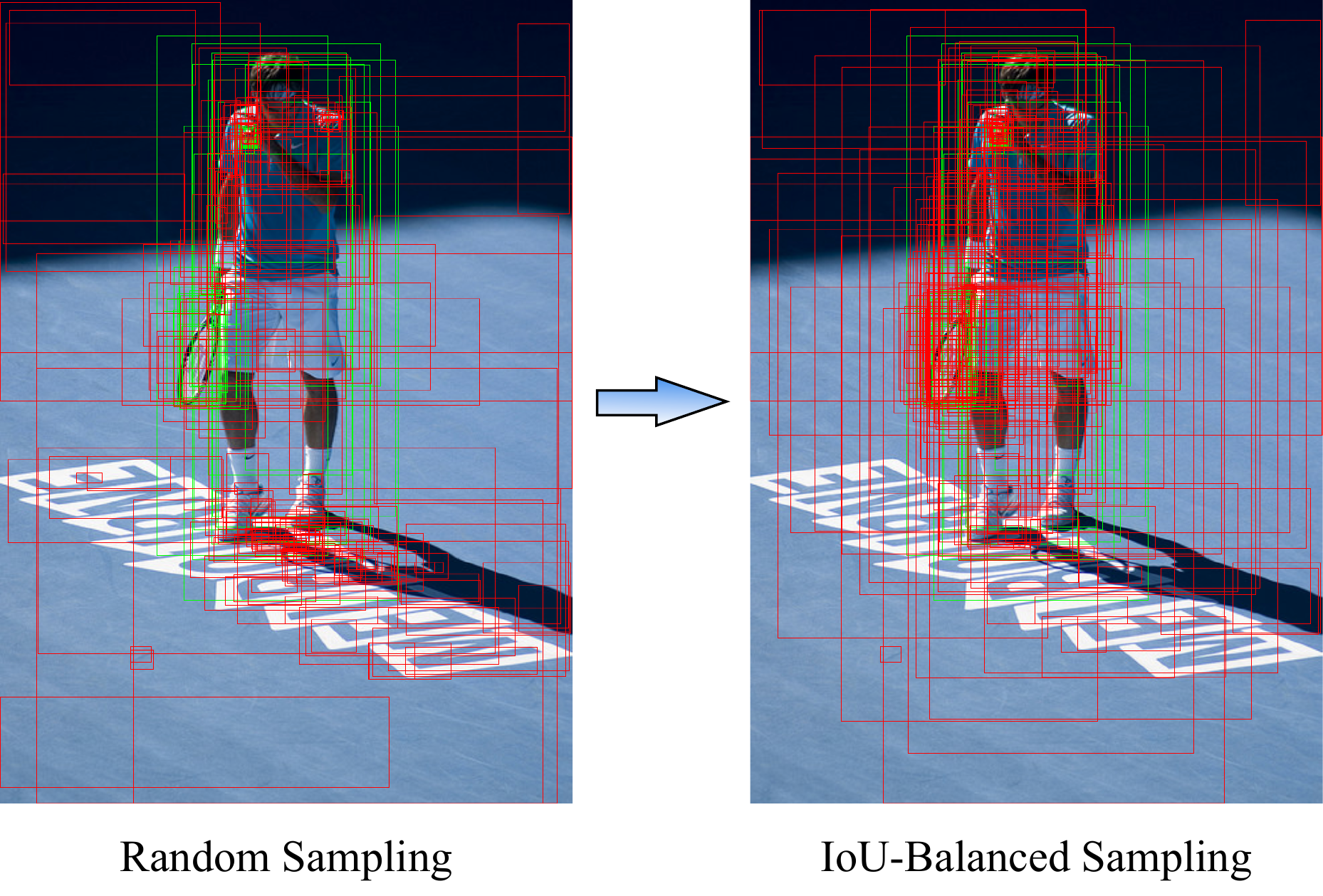}
	\caption{Visualization of training samples under random sampling and IoU-balanced sampling respectively.}
	\label{fig:sampling}
	\vspace{-10pt}
\end{figure}

\vspace{-12pt}
\paragraph{1) IoU-balanced Sampling.}
IoU-balanced sampling brings 0.9 points higher box AP than the ResNet-50 FPN Faster R-CNN baseline, validating the effectiveness of this cheap hard mining method.
We also visualize the training samples under random sampling and IoU-balanced sampling in Figure~\ref{fig:sampling}.
It can be seen that the selected samples are gathered to the regions where we are more interested in instead of randomly appearing around the target.

\vspace{-12pt}
\paragraph{2) Balanced Feature Pyramid.}
Balanced feature pyramid improves the box AP from 36.8 to 37.7.
Results in small, medium and large scales are consistently improved, which validate that the balanced semantic features balanced low-level and high-level information in each level and yield consistent improvements.

\vspace{-12pt}
\paragraph{3) Balanced L1 Loss.}
Balanced L1 loss improves the box AP from 37.7 to 38.5.
To be more specific, most of the improvements are from $AP_{75}$, which yields 1.1 points higher AP compared with corresponding baseline.
This result validates that the localization accuracy is much improved.

\vspace{-12pt}
\paragraph{Ablation Studies on IoU-balanced Sampling.}
Experimental results with different implementations of IoU-balanced sampling are shown in Table~\ref{tab:iou}.
We first verify the effectiveness of the complementary part, \ie sampling equal number of positive samples for each ground truth, which is stated in Section~\ref{subsec:sample} and denoted by \emph{Pos Balance} in Table~\ref{tab:iou}.
Since there are too little positive samples to explore the potential of this method, this sampling method provides only small improvements (0.2 points higher AP) compared to ResNet-50 FPN Faster R-CNN baseline.

Then we evaluate the effectiveness of IoU-balanced sampling for negative samples with different hyper-parameters $K$, which denotes the number of intervals.
Experiments in Table~\ref{tab:iou} show that the results are very close to each other when the parameter $K$ is set as 2, 3 or 5.
Therefore, the number of sampling interval is not much crucial in our IoU-balanced sampling, as long as the hard negatives are more likely selected.

\vspace{-12pt}
\paragraph{Ablation Studies on Balanced Feature Pyramid.}
Ablation studies of balanced feature pyramid are shown in Table~\ref{tab:fpn}.
We also report the experiments with PAFPN~\cite{panet}.
We first implement balanced feature pyramid only with integration.
Results show that the naive feature integration brings 0.4 points higher box AP than the corresponding baseline.
Note there is no refinement and no parameter added in this procedure.
With this simple method, each resolution obtains equal information from others.
Although this result is comparable with the one of PAFPN~\cite{panet}, we reach the feature level balance without extra convolutions, validating the effectiveness of this simple method.

Along with the embedded Gaussian non-local attention~\cite{nonlocal}, balanced feature pyramid can be further enhanced and improve the final results.
Our balanced feature pyramid is able to achieve 36.8 AP on COCO dataset, which is 0.9 points higher AP than ResNet-50 FPN Faster R-CNN baseline.
More importantly, the balanced semantic features have no conflict with PAFPN.
Based on the PAFPN, we include our feature balancing scheme and denote this implementation by Balanced PAFPN in Table~\ref{tab:fpn}.
Results show that the Balanced PAFPN is able to achieve 37.2 box AP on COCO dataset, with 0.9 points higher AP compared with the PAFPN.

\begin{table}[t]
	\centering
	\caption{Ablation studies of balanced semantic pyramid on COCO \emph{val-2017}.}
	\vspace{-5pt}
	\addtolength{\tabcolsep}{-2pt}
	\begin{tabular}{*{12}{c}}
		\toprule
		Settings          & AP   & $\text{AP}_{50}$ & $\text{AP}_{75}$ & $\text{AP}_{S}$ & $\text{AP}_{M}$ & $\text{AP}_{L}$ \\
		\midrule
		Baseline          & 35.9 & 58.0             & 38.4             & 21.2            & 39.5            & 46.4            \\
		\midrule
		Integration       & 36.3 & 58.8             & 38.8             & 21.2            & 40.1            & 46.3            \\
		Refinement        & 36.8 & 59.5             & 39.5             & 22.3            & 40.6            & 46.5            \\
		\midrule
		PAFPN\cite{panet} & 36.3 & 58.4             & 39.0             & 21.7            & 39.9            & 46.3            \\
		Balanced PAFPN    & 37.2 & 60.0             & 39.8             & 22.7            & 40.8            & 47.4            \\
		\bottomrule
	\end{tabular}
	\vspace{-10pt}
	\label{tab:fpn}
\end{table}

\begin{table}[htb]
	\centering
	\caption{Ablation studies of balanced L1 loss on COCO \emph{val-2017}. The numbers in the parentheses indicate the loss weight.}
	\vspace{-5pt}
	\addtolength{\tabcolsep}{-2pt}
	\begin{tabular}{*{12}{c}}
		\toprule
		Settings                        & AP   & $\text{AP}_{50}$ & $\text{AP}_{75}$ & $\text{AP}_{S}$ & $\text{AP}_{M}$ & $\text{AP}_{L}$ \\
		\midrule
		Baseline                        & 35.9 & 58.0             & 38.4             & 21.2            & 39.5            & 46.4            \\
		\midrule
		loss weight = 1.5               & 36.4 & 58.0             & 39.7             & 20.8            & 39.9            & 47.5            \\
		loss weight = 2.0               & 36.2 & 57.3             & 39.5             & 20.2            & 40.0            & 47.5            \\
		L1 Loss (1.0)                   & 36.4 & 57.4 & 39.1 & 21.0 & 39.7 &  47.9\\
		L1 Loss (1.5)                   & 36.6 & 57.2 & 39.8 & 20.2 & 40.0 & 48.2 \\
		L1 Loss (2.0)                   & 36.4 & 56.5 & 39.6 & 20.1 & 39.8 & 48.2 \\
		\midrule
		$\alpha = 0.2$,  $\gamma = 1.0$ & 36.7 & 58.1             & 39.5             & 21.4            & 40.4            & 47.4            \\
		$\alpha = 0.3$,  $\gamma = 1.0$ & 36.5 & 58.2             & 39.2             & 21.6            & 40.2            & 47.2            \\
		$\alpha = 0.5$,  $\gamma = 1.0$ & 36.5 & 58.2             & 39.2             & 21.5            & 39.9            & 47.2            \\
		\midrule
		$\alpha = 0.5$,  $\gamma = 1.5$ & 37.2 & 58.0             & 40.0             & 21.3            & 40.9            & 47.9            \\
		$\alpha = 0.5$,  $\gamma = 2.0$ & 37.0 & 58.0             & 40.0             & 21.2            & 40.8            & 47.6            \\
		\bottomrule
	\end{tabular}
	\vspace{-10pt}
	\label{tab:loss}
\end{table}

\vspace{-12pt}
\paragraph{Ablation Studies on Balanced L1 Loss.}
Ablation studies of balanced L1 loss are shown in Table~\ref{tab:loss}.
We observe that the localization loss is mostly half of the recognition loss.
Therefore, we first verify the performance when raising loss weight directly.
Results show that tuning loss weight only improves the result by 0.5 points.
And the result with a loss weight of 2.0 starts to drop down.
These results show that the outliers bring negative influence on the training process, and leave the potential of model architecture from being fully exploited.
We also conduct experiments with L1 loss for comparisons.
Experiments show that the results are inferior to ours.
Although the overall results are improved, the AP$_{50}$ and AP$_S$ drop obviously.

In order to compare with tuning loss weight directly, we first validate the effectiveness of balanced L1 loss when $\gamma = 1$.
Balanced L1 loss is able to bring 0.8 points higher AP than baseline.
With our best setting, balanced L1 loss finally achieves $37.2$ AP, which is 1.3 points higher than the ResNet-50 FPN Faster R-CNN baseline.
These experimental results validate that our balanced L1 achieves a more balanced training and makes the model better converged.


\section{Conclusion}
In this paper, we systematically revisit the training process of detectors,
and find the potential of model architectures is not fully exploited due to the imbalance issues existing in the training process.
Based on the observation, we propose Libra R-CNN to balance the imbalance through an overall balanced design.
With the help of the simple but effective components, \ie IoU-balanced sampling, balanced feature pyramid and balanced L1 loss, Libra R-CNN brings significant improvements on the challenging MS COCO dataset.
Extensive experiments show that Libra R-CNN well generalizes to various backbones for both two-stage detectors and single-stage detectors.

\vspace{-10pt}
\paragraph{Acknowledgement}
This work is partially supported by
the Science and Technology Plan of Zhejiang Province of China~(No.~2017C01033),
the Civilian Fundamental Research~(No.~D040301),
the Collaborative Research grant from SenseTime Group~(CUHK Agreement No.~TS1610626 \& No.~TS1712093),
and the General Research Fund~(GRF) of Hong Kong~(No.~14236516 \& No.~14203518).

{\small
\bibliographystyle{ieee_fullname}
\bibliography{egbib}

\begin{thebibliography}{10}\itemsep=-1pt

\bibitem{bell2016inside}
Sean Bell, C Lawrence~Zitnick, Kavita Bala, and Ross Girshick.
\newblock Inside-outside net: Detecting objects in context with skip pooling
  and recurrent neural networks.
\newblock In {\em IEEE Conference on Computer Vision and Pattern Recognition},
  2016.

\bibitem{cai2016unified}
Zhaowei Cai, Quanfu Fan, Rogerio~S Feris, and Nuno Vasconcelos.
\newblock A unified multi-scale deep convolutional neural network for fast
  object detection.
\newblock In {\em European Conference on Computer Vision}, 2016.

\bibitem{cascadercnn}
Zhaowei Cai and Nuno Vasconcelos.
\newblock Cascade r-cnn: Delving into high quality object detection.
\newblock In {\em IEEE Conference on Computer Vision and Pattern Recognition},
  2018.

\bibitem{htc}
Kai Chen, Jiangmiao Pang, Jiaqi Wang, Yu Xiong, Xiaoxiao Li, Shuyang Sun,
  Wansen Feng, Ziwei Liu, Jianping Shi, Wanli Ouyang, and Dahua Lin.
\newblock Hybrid task cascade for instance segmentation.
\newblock {\em arXiv preprint arXiv:1901.07518}, 2019.

\bibitem{mmdetection2018}
Kai Chen, Jiangmiao Pang, Jiaqi Wang, Yu Xiong, Xiaoxiao Li, Shuyang Sun,
  Wansen Feng, Ziwei Liu, Jianping Shi, Wanli Ouyang, Chen~Change Loy, and
  Dahua Lin.
\newblock mmdetection.
\newblock \url{https://github.com/open-mmlab/mmdetection}, 2018.

\bibitem{rfcn}
Jifeng Dai, Yi Li, Kaiming He, and Jian Sun.
\newblock R-fcn: Object detection via region-based fully convolutional
  networks.
\newblock In {\em Advances in Neural Information Processing Systems}, 2016.

\bibitem{fastrcnn}
Ross Girshick.
\newblock Fast r-cnn.
\newblock In {\em IEEE Conference on Computer Vision and Pattern Recognition},
  2015.

\bibitem{rcnn}
Ross Girshick, Jeff Donahue, Trevor Darrell, and Jitendra Malik.
\newblock Rich feature hierarchies for accurate object detection and semantic
  segmentation.
\newblock In {\em IEEE Conference on Computer Vision and Pattern Recognition},
  2014.

\bibitem{Detectron2018}
Ross Girshick, Ilija Radosavovic, Georgia Gkioxari, Piotr Doll\'{a}r, and
  Kaiming He.
\newblock Detectron.
\newblock \url{https://github.com/facebookresearch/detectron}, 2018.

\bibitem{maskrcnn}
Kaiming He, Georgia Gkioxari, Piotr Doll{\'a}r, and Ross Girshick.
\newblock Mask r-cnn.
\newblock In {\em IEEE International Conference on Computer Vision}, 2017.

\bibitem{spp}
Kaiming He, Xiangyu Zhang, Shaoqing Ren, and Jian Sun.
\newblock Spatial pyramid pooling in deep convolutional networks for visual
  recognition.
\newblock In {\em European Conference on Computer Vision}, 2014.

\bibitem{resnet}
Kaiming He, Xiangyu Zhang, Shaoqing Ren, and Jian Sun.
\newblock Deep residual learning for image recognition.
\newblock In {\em IEEE Conference on Computer Vision and Pattern Recognition},
  2016.

\bibitem{learningnms}
Jan~Hendrik Hosang, Rodrigo Benenson, and Bernt Schiele.
\newblock Learning non-maximum suppression.
\newblock In {\em IEEE Conference on Computer Vision and Pattern Recognition},
  2017.

\bibitem{relationnetwork}
Han Hu, Jiayuan Gu, Zheng Zhang, Jifeng Dai, and Yichen Wei.
\newblock Relation networks for object detection.
\newblock In {\em IEEE Conference on Computer Vision and Pattern Recognition},
  2018.

\bibitem{iounet}
Borui Jiang, Ruixuan Luo, Jiayuan Mao, Tete Xiao, and Yuning Jiang.
\newblock Acquisition of localization confidence for accurate object detection.
\newblock {\em arXiv preprint arXiv:1807.11590}, 1, 2018.

\bibitem{kendall2017multi}
Alex Kendall, Yarin Gal, and Roberto Cipolla.
\newblock Multi-task learning using uncertainty to weigh losses for scene
  geometry and semantics.
\newblock {\em arXiv preprint arXiv:1705.07115}, 3, 2017.

\bibitem{kong2018deep}
Tao Kong, Fuchun Sun, Wenbing Huang, and Huaping Liu.
\newblock Deep feature pyramid reconfiguration for object detection.
\newblock {\em arXiv preprint arXiv:1808.07993}, 2018.

\bibitem{cornernet}
Hei Law and Jia Deng.
\newblock Cornernet: Detecting objects as paired keypoints.
\newblock In {\em European Conference on Computer Vision}, 2018.

\bibitem{fpn}
Tsung-Yi Lin, Piotr Doll{\'a}r, Ross~B Girshick, Kaiming He, Bharath Hariharan,
  and Serge~J Belongie.
\newblock Feature pyramid networks for object detection.
\newblock In {\em IEEE Conference on Computer Vision and Pattern Recognition},
  2017.

\bibitem{focalloss}
Tsung-Yi Lin, Priyal Goyal, Ross Girshick, Kaiming He, and Piotr Doll{\'a}r.
\newblock Focal loss for dense object detection.
\newblock {\em IEEE Transactions on Pattern Analysis and Machine Intelligence},
  2018.

\bibitem{coco}
Tsung-Yi Lin, Michael Maire, Serge Belongie, James Hays, Pietro Perona, Deva
  Ramanan, Piotr Doll{\'a}r, and C~Lawrence Zitnick.
\newblock Microsoft coco: Common objects in context.
\newblock In {\em European Conference on Computer Vision}, 2014.

\bibitem{panet}
Shu Liu, Lu Qi, Haifang Qin, Jianping Shi, and Jiaya Jia.
\newblock Path aggregation network for instance segmentation.
\newblock In {\em IEEE Conference on Computer Vision and Pattern Recognition},
  2018.

\bibitem{ssd}
Wei Liu, Dragomir Anguelov, Dumitru Erhan, Christian Szegedy, Scott Reed,
  Cheng-Yang Fu, and Alexander~C Berg.
\newblock Ssd: Single shot multibox detector.
\newblock In {\em European Conference on Computer Vision}, 2016.

\bibitem{ouyang2017chained}
Wanli Ouyang, Kun Wang, Xin Zhu, and Xiaogang Wang.
\newblock Chained cascade network for object detection.
\newblock In {\em IEEE International Conference on Computer Vision}, 2017.

\bibitem{pytorch}
Adam Paszke, Sam Gross, Soumith Chintala, Gregory Chanan, Edward Yang, Zachary
  DeVito, Zeming Lin, Alban Desmaison, Luca Antiga, and Adam Lerer.
\newblock Automatic differentiation in pytorch.
\newblock 2017.

\bibitem{yolo}
Joseph Redmon, Santosh Divvala, Ross Girshick, and Ali Farhadi.
\newblock You only look once: Unified, real-time object detection.
\newblock In {\em IEEE Conference on Computer Vision and Pattern Recognition},
  2016.

\bibitem{yolo9000}
Joseph Redmon and Ali Farhadi.
\newblock Yolo9000: better, faster, stronger.
\newblock {\em arXiv preprint}, 2017.

\bibitem{frcnn}
Shaoqing Ren, Kaiming He, Ross Girshick, and Jian Sun.
\newblock Faster r-cnn: Towards real-time object detection with region proposal
  networks.
\newblock In {\em Advances in Neural Information Processing Systems}, 2015.

\bibitem{ohem}
Abhinav Shrivastava, Abhinav Gupta, and Ross Girshick.
\newblock Training region-based object detectors with online hard example
  mining.
\newblock In {\em IEEE Conference on Computer Vision and Pattern Recognition},
  2016.

\bibitem{snip}
Bharat Singh and Larry~S Davis.
\newblock An analysis of scale invariance in object detection--snip.
\newblock In {\em IEEE Conference on Computer Vision and Pattern Recognition},
  2018.

\bibitem{sniper2018}
Bharat Singh, Mahyar Najibi, and Larry~S Davis.
\newblock {SNIPER}: Efficient multi-scale training.
\newblock {\em NIPS}, 2018.

\bibitem{nonlocal}
Xiaolong Wang, Ross Girshick, Abhinav Gupta, and Kaiming He.
\newblock Non-local neural networks.
\newblock {\em arXiv preprint arXiv:1711.07971}, 10, 2017.

\bibitem{resnext}
Saining Xie, Ross Girshick, Piotr Doll{\'a}r, Zhuowen Tu, and Kaiming He.
\newblock Aggregated residual transformations for deep neural networks.
\newblock In {\em IEEE Conference on Computer Vision and Pattern Recognition},
  2017.

\bibitem{unitbox}
Jiahui Yu, Yuning Jiang, Zhangyang Wang, Zhimin Cao, and Thomas Huang.
\newblock Unitbox: An advanced object detection network.
\newblock In {\em Proceedings of the 24th ACM international conference on
  Multimedia}, pages 516--520. ACM, 2016.

\bibitem{zeiler2014visualizing}
Matthew~D Zeiler and Rob Fergus.
\newblock Visualizing and understanding convolutional networks.
\newblock In {\em European Conference on Computer Vision}, 2014.

\bibitem{zeng2018crafting}
Xingyu Zeng, Wanli Ouyang, Junjie Yan, Hongsheng Li, Tong Xiao, Kun Wang, Yu
  Liu, Yucong Zhou, Bin Yang, Zhe Wang, et~al.
\newblock Crafting gbd-net for object detection.
\newblock {\em IEEE transactions on pattern analysis and machine intelligence},
  40(9):2109--2123, 2018.

\bibitem{adapt}
Xinge Zhu, Jiangmiao Pang, Ceyuan Yang, Jianping Shi, and Dahua Lin.
\newblock Adapting object detectors via selective cross-domain alignment.
\newblock In {\em IEEE Conference on Computer Vision and Pattern Recognition},
  2019.

\end{thebibliography}
}

\end{document}